\documentclass[11pt,a4paper]{article}
\usepackage[hyperref]{acl2020}
\usepackage{times}
\usepackage{latexsym}

\usepackage{microtype}

\usepackage{float}
\usepackage{color}
\usepackage{xcolor}

\usepackage{booktabs}
\usepackage{url}
\usepackage{graphicx}
\usepackage{amsmath,amssymb}

\usepackage{gb4e}
\noautomath

\aclfinalcopy 
\def\aclpaperid{750} 

\newcommand{\figref}[1]{Figure~\ref{#1}}

\newcommand{\sectionref}[1]{Section~\ref{#1}}

\definecolor{Orange}{RGB}{255,140,0}

\definecolor{Pink}{RGB}{150,0,100}

\definecolor{Green}{RGB}{0,150,0}

\newcommand{\prtt}[1]{\textcolor{blue}{(#1)}}
\newcommand{\prmod}[1]{\textcolor{orange}{(#1)}}
\newcommand{\pomod}[1]{\textcolor{olive}{(#1)}}

\title{Harnessing the linguistic signal to predict scalar inferences}
 
\author{
  Sebastian Schuster\thanks{\ \ \ Equal contribution.} \\
  Stanford University \\
   {\small \texttt{sebschu@stanford.edu}}   \\\And
  Yuxing Chen${}^*$ \\
  Stanford University \\ 
  {\small \texttt{yxchen28@stanford.edu}} \\\And
  Judith Degen \\
  Stanford University \\
  {\small \texttt{jdegen@stanford.edu}}}

\begin{document}

\maketitle

\begin{abstract}

Pragmatic inferences often subtly depend on the presence or absence of linguistic features. For example, the presence of a partitive construction  (\emph{of the}) increases the strength of a so-called scalar inference: listeners perceive the inference that Chris did not eat all of the cookies to be stronger after hearing \emph{``Chris ate some of the cookies''} than after hearing the same utterance without a partitive, \emph{``Chris ate some cookies''}. In this work, we explore to what extent neural network sentence encoders can learn to predict the strength of scalar inferences. We first show that an LSTM-based sentence encoder trained on an English dataset of human inference strength ratings is able to predict ratings with high accuracy ($r=0.78$). We then probe the model's behavior using manually constructed minimal sentence pairs and corpus data. We find that the model inferred previously established associations between linguistic features and inference strength, suggesting that the model learns to use linguistic features to predict pragmatic inferences.

\end{abstract}

\section{Introduction}

An important property of human communication is that listeners can {infer} information beyond the literal meaning of an utterance. One well-studied type of inference is \textit{scalar inference} \citep{Grice1975,Horn1984}, whereby a listener who hears an utterance with a scalar item like \textit{some} infers the negation of a stronger alternative with \textit{all}:

\begin{exe}
\small
\ex
\begin{xlist}
\ex \label{ex:cookie} Chris ate some of the cookies.
\ex \label{ex:cookie-inf} $\rightsquigarrow$ Chris ate some, but not all, of the cookies. 
\end{xlist}
\end{exe}

Early accounts of scalar inferences (e.g., \citealt{gazdar1979pragmatics,Horn1984,levinson2000presumptive}) considered them to arise by default unless explicitly contradicted in context. However, in a recent corpus study, \newcite{degen2015investigating} showed that there is much more variability in scalar inferences from \textit{some} to \emph{not all} than previously assumed. \newcite{degen2015investigating} further showed that this variability is not random and that several lexical, syntactic, and semantic/pragmatic features of context explain much of the variance in inference strength.\footnote{See Section~\ref{sec:dataset} for the operationalization of inference strength that we use throughout this paper and for a description of these features.}

Recent Bayesian game-theoretic models of pragmatic reasoning \citep{goodman2016pragmatic,franke2016probabilistic} are able to integrate speaker expectations with world knowledge to predict listeners' pragmatic inferences in many cases (e.g., \citealt{goodman2013knowledge,degen2015wonky}). However, to compute speaker expectations, these models require manual specification of features as well as specification of a finite set of possible utterances. Further, inference becomes intractable when scaling up beyond toy domains to make predictions for arbitrary utterances.\footnote{Recent models of generating pragmatic image descriptions \citep{Andreas2016,CohnGordon2018} and color descriptions \citep{Monroe2017} have overcome this issue by approximating the distributions of utterances given a set of potential referents. However, these models require a finite set of world states (e.g., several referents to choose from) and a corresponding generative model of utterances (e.g., an image captioning model) and are therefore also limited to scenarios with pre-specified world states and a corresponding generative model.} Neural network (NN) models, on the other hand, do not suffer from these limitations: they are capable of making predictions for arbitrary utterances and do not require manual specification of features. Unlike Bayesian game-theoretic models, however, NN models have no explicit pragmatic reasoning mechanisms.

In this work, we investigate to what extent NN models can learn to predict subtle differences in scalar inferences and to what extent these models infer associations between linguistic features and inference strength. 
In this enterprise we follow  the recent successes of  NN models in predicting a range of linguistic phenomena such as long distance syntactic dependencies (e.g., \citealt{Elman1990,Linzen2016,Gulordava2018,Futrell2019,Wilcox2019}), semantic entailments (e.g., \citealt{Bowman2015,Conneau2018}), acceptability judgements \citep{Warstadt2018}, factuality \citep{Rudinger2018},  negative polarity item licensing environments \citep{Warstadt2019}, and,  to  some  extent, speaker commitment \citep{Jiang2019}.  In particular, we ask: 
\vspace{-.4em}
\begin{enumerate}
\itemsep0em 
	\item How well can a neural network sentence encoder learn to predict human inference strength judgments for utterances with \emph{some}? 
	\item To what extent does such a model capture the qualitative effects of hand-mined contextual features previously identified as influencing inference strength?
\end{enumerate}
\vspace{-.4em}
To address the first question, we compare the performance of several NN models that differ in the underlying word embedding model (GloVe, ELMo, or BERT). To address the second question, we probe the best model's behavior through an analysis of predictions on manually constructed minimal sentence pairs, a regression analysis, and an analysis of attention weights. We find that the best model is able to predict inference strength ratings on a held-out test set with high accuracy ($r=0.78$). The three analyses consistently suggest that the model learned associations between inference strength and linguistic features established by previous work \citep{degen2015investigating}.

We release data and code at \url{https://github.com/yuxingch/Implicature-Strength-Some}.

\section{The dataset} \label{sec:dataset}

We use the annotated dataset collected by \newcite{degen2015investigating}, a dataset of the utterances from the Switchboard corpus of English telephone dialogues \citep{godfrey1992switchboard} with a noun phrase (NP) with \emph{some}.  The dataset consists of 1,362 unique utterances. For each example with a \emph{some}-NP, \newcite{degen2015investigating} collected inference strength ratings from at least 10 participants recruited on Amazon's Mechanical Turk. Participants saw both the target utterance and ten utterances from the preceding discourse context. They then rated the similarity between the original utterance like (\ref{utteranceA}) and an utterance in which \emph{some} was replaced with \emph{some, but not all} like (\ref{utteranceB}), on a 7-point Likert scale with endpoints labeled ``very different meaning'' (1) and ``same meaning'' (7). Low similarity ratings thus indicate low inference strength, and high similarity ratings indicate high inference strength.
\vspace{-.2em}
\begin{exe}
\small
    \ex \begin{xlist}
        \ex \label{utteranceA}I like -- I like to read \emph{some} of the philosophy stuff. 
        \ex \label{utteranceB}I like -- I like to read \emph{some, but not all,} of the philosophy stuff.
    \end{xlist}
\end{exe}
\vspace{-.4em}

Using this corpus, \newcite{degen2015investigating} found that several linguistic and contextual factors influenced inference strength ratings, including the partitive form \emph{of}, subjecthood, the previous mention of the  NP referent, determiner strength, and modification of the head noun, which we describe in turn.

\noindent\textbf{Partitive:}
(\ref{ex:partitive}a-b) are example utterances from the corpus with and without partitive \emph{some}-NPs, respectively. Values in parentheses indicate the mean inference strength rating for that item. On average, utterances with partitives yielded stronger inference ratings than ones without.
\vspace{-.2em}
\begin{exe}
\small
    \ex \label{ex:partitive}  
    \begin{xlist}
    \ex We [...] buy \textit{some} of our own equipment. \hfill (5.3)
    \ex  You sound like you have \emph{some} small ones in the background. \hfill (1.5)
    \end{xlist}
\end{exe}
\vspace{-.4em}

\noindent\textbf{Subjecthood:}
Utterances in which the \emph{some}-NP appears in subject position, as in (\ref{ex:subj}a), yielded stronger inference ratings than utterances in which the \emph{some}-NP appears in a different grammatical position, e.g., as a direct object as in  (\ref{ex:subj}b).
\vspace{-.2em}
\begin{exe}
\small
    \ex \label{ex:subj} 
    \begin{xlist}
    \ex \emph{Some} kids are really having it. \hfill (5.9)
    \ex That would take \emph{some} planning. \hfill (1.4)
    \end{xlist}
\end{exe}
\vspace{-.4em}
\noindent\textbf{Previous mention:}
Discourse properties also have an effect on inference strength. A \emph{some}-NP with a previously mentioned embedded NP referent yields stronger inferences than a \emph{some}-NP whose embedded NP referent has not been previously mentioned. For example, (\ref{ex:mention}a) contains a \emph{some}-NP in which \emph{them} refers to previously mentioned \emph{Mission Impossible tape recordings}, whereas \emph{problems} in the \emph{some}-NP in (\ref{ex:mention}b) has not been previously mentioned.
\vspace{-.2em}
\begin{exe}
\small
    \ex \label{ex:mention}\begin{xlist}
    \ex  I've seen \emph{some} of them on repeats. \hfill (5.8)
    \ex  What do you feel  are \emph{some} of the main problems? \hfill (3.4)  
    \end{xlist}
\end{exe}
\vspace{-.4em}

\noindent\textbf{Modification:} \citet{degen2015investigating} also found a small effect of whether or not the head noun of the \emph{some}-NP was modified: \emph{some}-NPs with unmodified head nouns yielded slightly stronger inferences than those with modified head nouns.

\noindent\textbf{Determiner strength:} Finally, it has been argued that there are two types of \emph{some}: a weak \emph{some} and a strong \emph{some} \cite{milsark1974,barwise1981}. This weak/strong distinction has been notoriously hard to pin down \cite{horn1997} and \newcite{degen2015investigating} used empirical strength norms elicited independently for each item. To this end, she exploited the fact that removing weak \textit{some} from an utterance has little effect on its meaning whereas removing strong \emph{some} changes the meaning.  Determiner strength ratings were thus elicited by asking participants to rate the similarity between the original utterance and an utterance without \emph{some (of)} on a 7-point Likert scale from `different meaning' to `same meaning'. Items with stronger \emph{some} -- e.g., (\ref{ex:strength}a), determiner strength 3.3 -- yielded stronger inference ratings than items with weaker \emph{some} -- e.g., (\ref{ex:strength}b), determiner strength 6.7.
\vspace{-.2em}
\begin{exe}
\small
    \ex \label{ex:strength} \begin{xlist}
    \ex And \emph{some} people don't vote. \hfill (5.2)
    \ex  Well, we could use \emph{some} rain up here. \hfill (2.1)
    \end{xlist}
\end{exe}
\vspace{-.4em}

The quantitative findings from \newcite{degen2015investigating} are summarized in \figref{fig:cv-coefficients}, which shows in blue the regression coefficients for all predictors she considered (see the original paper for more detailed descriptions). 

For our experiments, we randomly split the dataset into a 70\% training and 30\% test set, resulting in 954 training items and 408 test items.

\section{Model}

\begin{figure}[t]
	\includegraphics[width=.48\textwidth]{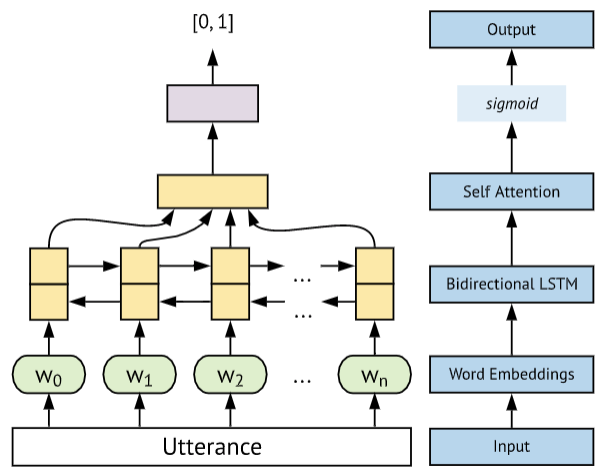}
	\caption{Model architecture.}
		\label{fig:model-architecture}
\end{figure}

The objective of the model is to predict mean inference strength rating $i$ given an utterance (a sequence of words) $U = \{w_1,w_2,...,w_N\}$. We rescale the 1-to-7 Likert scale ratings to the interval $[0,1]$. Figure~\ref{fig:model-architecture} shows the overall model architecture. The model is a sentence classification model akin to the model proposed by \citet{lin2017structured}. It first embeds the utterance tokens using  pre-trained embedding models, and then forms a sentence representation by passing the embedded tokens through a 2-layer bidirectional LSTM network (biLSTM) \cite{hochreiter1997long} with dropout \cite{srivastava2014dropout} followed by a self-attention mechanism that provides a weighted average of the hidden states of the top-most biLSTM layer. This sentence representation is then passed through a transformation layer with a sigmoid activation function, which outputs the predicted score in the interval $[0,1]$.

\section{Experiments}

\subsection{Training}

We used 5-fold cross-validation on the training data to optimize the following hyperparameters.

\noindent \textbf{Word embedding model}: 100d GloVe \cite{pennington2014glove}, 1024d ELMo \cite{peters2018deep,gardner2018allennlp}, 768d BERT-base, 1024d BERT-large \cite{devlin2019bert,wolf2019transformers}.\\
\textbf{Output layer of word embedding models}: $[1,3]$ for ELMo, $[1,12]$ for BERT-base, and $[1,24]$ for BERT-large.\\
\textbf{Dimension of LSTM hidden states}: $\{100,200,400,800\}$.\\
\textbf{Dropout rate in LSTM}: $\{0.1,0.2,0.3,0.4\}$. 

We first optimized the output layer parameter for each contextual word embedding model while keeping all other parameters fixed. We then optimized the other parameters for each embedding model by computing the average correlation between the model predictions and the human ratings across the five cross-validation folds.

\noindent \textbf{Architectural variants.} We also evaluated all combinations of two architectural variants: First, we evaluated models in which we included the attention layer (\textsc{LSTM+Attention}) or simply used the final hidden state of the LSTM (\textsc{LSTM}) as a sentence representation. Second, since participants providing inference strength ratings also had access to 10 utterances from the  preceding conversational context, we also compared models that make predictions based only the target utterance with the \emph{some}-NP and models that make predictions based on target utterances and the preceding conversational context. 
For the models using GloVe and ELMo, we prepended the conversational context to the target utterance to obtain a joint context and utterance embedding. For models using BERT, we made use of the fact that BERT had been trained to jointly embed two sentences or documents, and we obtained embeddings for the tokens in the target utterance by feeding the target utterance as the first document and the preceding context as the second document into the BERT encoder. 
We discarded the hidden states of the preceding context and only used the output of BERT for the tokens in the target utterance.

\noindent \textbf{Implementation details.} We implemented the model in PyTorch \cite{paszke2017automatic}. We trained the model using the Adam optimizer \cite{kingma2014adam} with default parameters and a learning rate of 0.001, minimizing the mean squared error of the predicted ratings. 
In the no-context experiments, we truncated target utterances longer than 30 tokens, and in the experiments with context, we truncated the beginning of the preceding context such that the number of tokens did not exceed 150.

\noindent \textbf{Evaluation.} We evaluated the model predictions in terms of their correlation $r$ with the human inference strength ratings.
As mentioned above, we optimized the hyperparameters using cross validation. We then took 
the best set of parameters and trained a model on all the available training data and evaluated that model on the held-out data.

\subsection{Tuning results}

Not surprisngly, we find that the attention layer improves predictions and that contextual word embeddings lead to better results than the static GloVe
embeddings. We also find that including the conversational context does not improve predictions (see Appendix~\ref{app:hyperparameter}, for learning curves of all models, and Section~\ref{sec:context}, for a discussion of the role of conversational context).

Otherwise, the model is quite insensitive to hyperparameter settings: neither the dimension of the hidden LSTM states nor the dropout rate had considerable effects
on the prediction accuracy. We do find, however, that there are differences depending on the BERT and ELMo layer that we use as word representations.
We find that higher layers work better than lower layers, suggesting that word representations that are influenced by other utterance tokens are helpful for this 
task.

Based on these optimization runs, we chose the model with attention that uses the BERT-large embeddings but no conversational context for the subsequent experiments and analyses.

\subsection{Test results}
\begin{figure}[t]
\center
\includegraphics[width=0.75\columnwidth]{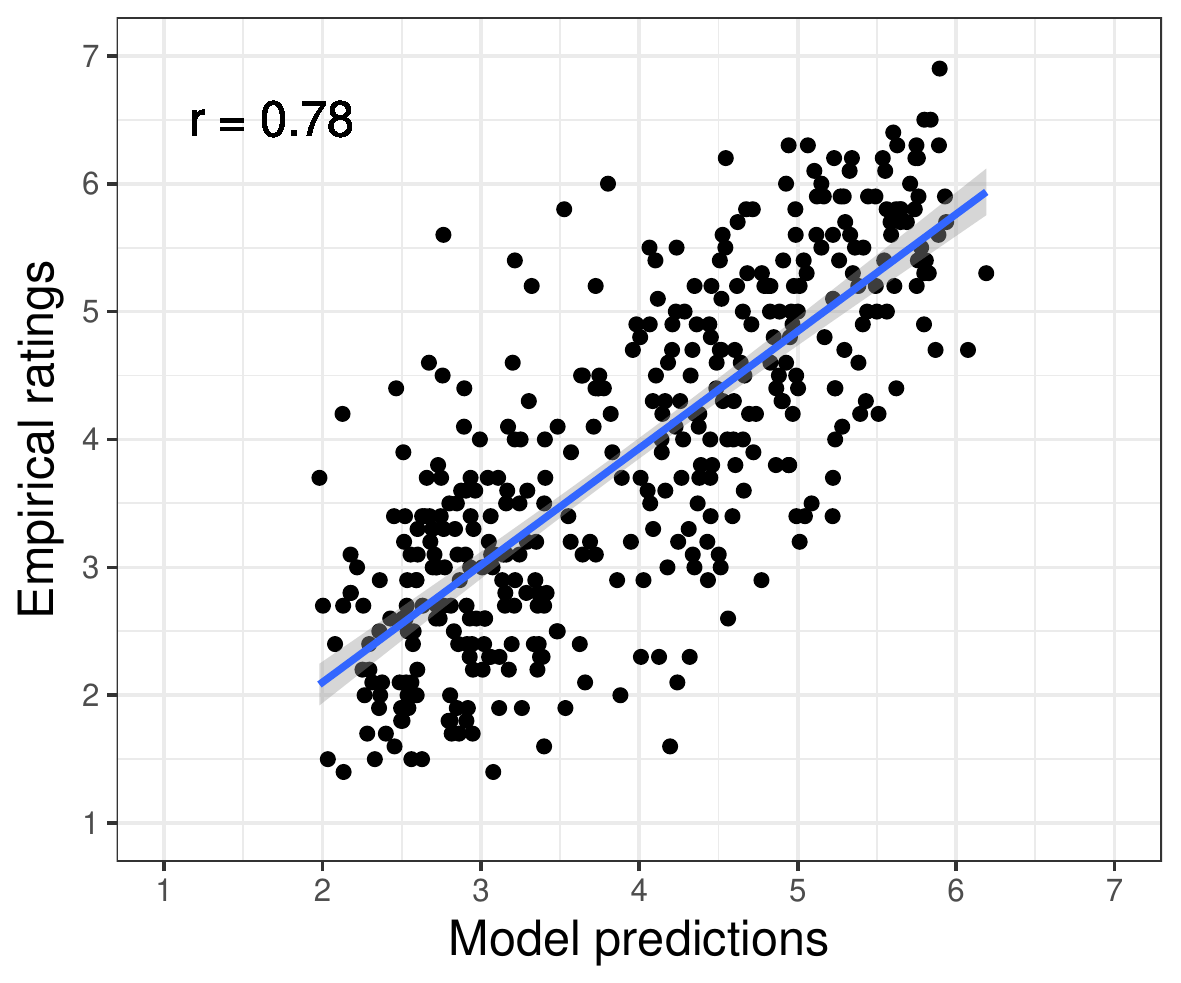}
\caption{Correlation between empirical ratings and predictions of the \textsc{BERT-large LSTM+Attention} model on held-out test items.}
	\label{fig:eval-correlations}
\end{figure}

\figref{fig:eval-correlations} shows the correlation between the best model according to the tuning runs (now trained on all training data) and the empirical ratings on the 408 held-out test items. As this plot shows, the model predictions fall within a close range of the empirical ratings for most of the items ($r=0.78$).\footnote{For comparison, we estimated how well the human ratings correlated through a bootstrapping analysis: We re-sampled the human ratings for each item and computed the average correlation coefficient between the original and the re-sampled datasets, which we found to be approximately 0.93.} Further, similarly as in the empirical data, there seem to be two clusters in the model predictions: one that includes lower ratings and one that includes higher ratings, corresponding to strong and weak scalar inferences, respectively. The only systematic deviation appears to be that the model does not predict any extreme ratings -- almost all predictions are greater than 2 or less than 6, whereas the empirical ratings include some cases outside of this range. 

Overall, these results suggest that the model can learn to closely predict the strength of scalar inferences.  However, this result by itself does not provide evidence that the model learned associations between linguistic features and inference strength, since it could also be that, given the large number of parameters, the model learned spurious correlations independent of the empirically established feature-strength associations. To investigate whether the model learned the expected associations, we probed the model's behavior in multiple ways, which we discuss next.

\section{Model behavior analyses}
\label{sec:analyses}

\paragraph{Minimal pair analysis.} As a first analysis, we constructed artificial minimal pairs that differed along 
several factors of interest and compared the model predictions. Such methods have been recently used to probe, for example, what kind of syntactic dependencies different types of recurrent neural network language models are capable of encoding or to what extent sentence vector representations capture compositional meanings (e.g., \citealt{Linzen2016, Gulordava2018, chowdhury2018rnn, Ettinger2018,marvin2018targeted,Futrell2019, Wilcox2019}), and also allow us to probe whether the model is sensitive to controlled changes in the input.

We constructed a set of 25 initial sentences with \emph{some}-NPs. For each sentence, we created 32 variants that differed in the following four properties of the \emph{some}-NP: subjecthood, \textcolor{blue}{partitive}, \textcolor{orange}{pre-nominal modification}, and \textcolor{olive}{post-nominal modification}. For the latter three features, we either included or excluded \emph{of the} or the modifier, respectively. For example, the version in  (\ref{ex:farmersa}a) includes \emph{of the} whereas the version in  (\ref{ex:farmersa}b)  lacks the partitive feature. To manipulate subjecthood of the \emph{some}-NP, we created variants in which \emph{some} was either the determiner in the subject NP as in (\ref{ex:farmersa}) or in the object-NP as in (\ref{ex:farmersb}). We also created passive versions of each of these variants (\ref{ex:farmersc}-\ref{ex:farmersd}). Each set of sentences included
a unique main verb, a unique pair of NPs, and unique modifiers.
The full list of sentences can be found in Appendix~\ref{app:sentences}.

\begin{exe}
\small
        \ex     \label{ex:farmersa}
          \begin{xlist} \ex Some \textcolor{blue}{of the} \prmod{organic} farmers \pomod{in the mountains} milked the brown goats who graze on the meadows.
          \ex Some \prmod{organic} farmers \pomod{in the mountains} milked the brown goats who graze on the meadows.
            \end{xlist}
        
        \ex  \label{ex:farmersb} The organic farmers in the mountains milked some \prtt{of the} \prmod{brown} goats \pomod{who graze on the meadows}.
        \ex \label{ex:farmersc} The brown goats who graze on the meadows were milked by some \prtt{of the} \prmod{organic} farmers \pomod{in the mountains}.
        \ex \label{ex:farmersd} Some \prtt{of the} \prmod{brown} goats \pomod{who graze on the meadows} were milked by the organic farmers in the mountains.

\end{exe}

\begin{figure}[t]
	\includegraphics[width=.49\textwidth]{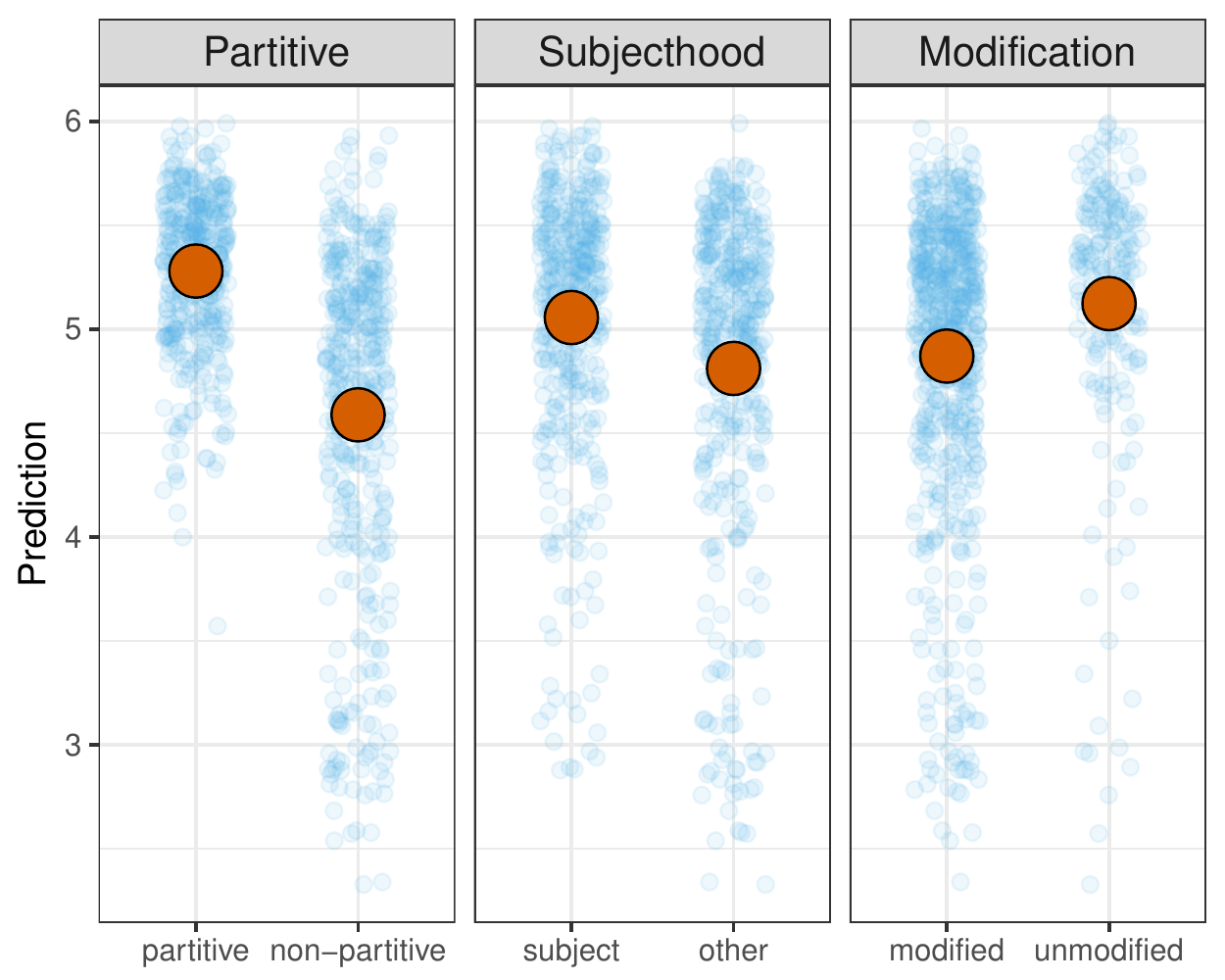}
	\caption{Average model predictions on manually constructed sentences, grouped by presence of partitives, by grammatical function of the \emph{some}-NP, and by presence of nominal modifiers.  Semi-transparent dots show predictions on individual sentences.}
	\label{fig:minimal-pairs}
\end{figure}

\figref{fig:minimal-pairs} shows the model predictions for the manually constructed sentences grouped by the presence of a partitive construction, the grammatical function of the \emph{some}-NP, and the presence of a modifier. As in the natural dataset from \citet{degen2015investigating}, sentences with a partitive received higher predicted ratings than sentences without a partitive; sentences with subject \emph{some}-NPs received higher predicted ratings than sentences with non-subject \emph{some}-NPs; and sentences with a modified head noun in the \emph{some}-NP received lower predictions than sentences with an unmodified \emph{some}-NP. All these results provide evidence that the model learned the correct associations. This is particularly remarkable considering the train-test mismatch: the model was trained on noisy transcripts of spoken language that contained many disfluencies and repairs, and was subsequently tested on clean written sentences.    

\begin{figure*}[t]
\center
\includegraphics[width=\textwidth,trim={0 0.4cm 0 0},clip,width=\textwidth]{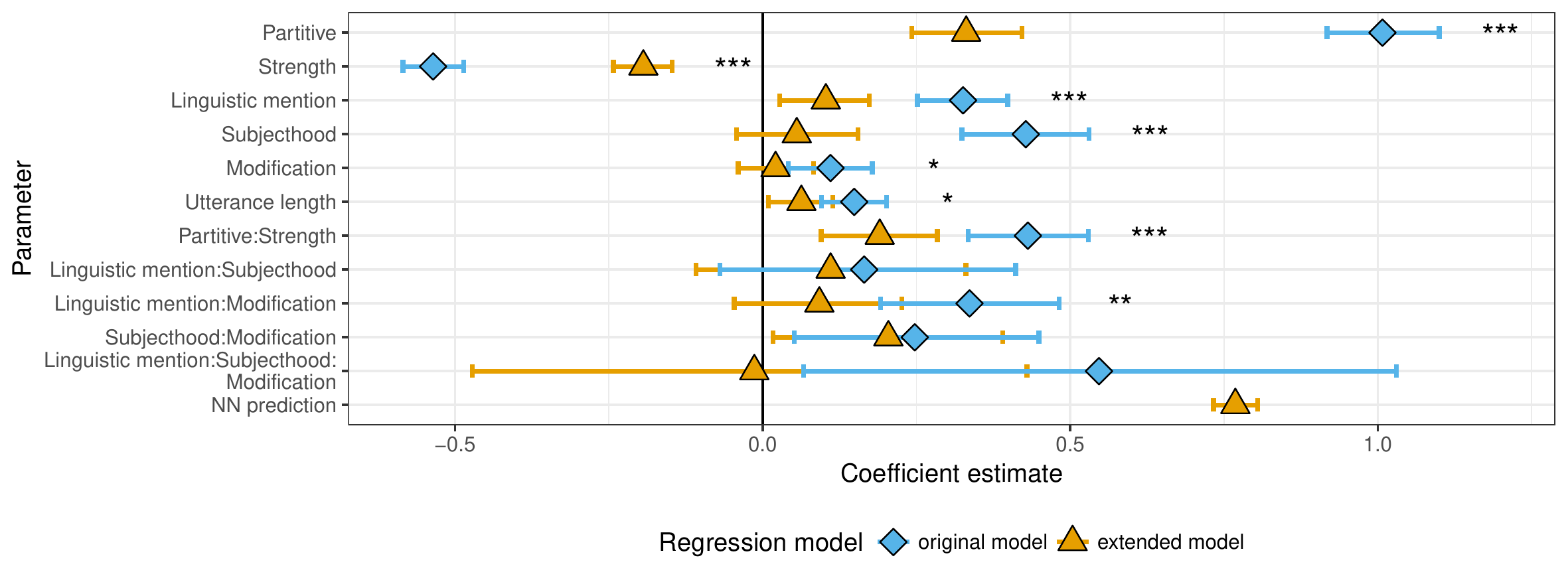}
\caption{Maximum a posteriori estimates and 95\%-credible intervals of coefficients for original and extended Bayesian mixed-effects regression models predicting the inference strength ratings. */**/*** indicate that the probability of the coefficient of the original model having a larger magnitude than the coefficient of the extended model is less than 0.05, 0.01, and 0.001, respectively.}
	\label{fig:cv-coefficients}
\end{figure*} 

\textbf{Regression analysis.} In the minimal pair analysis above we only investigated model predictions for three factors. As a second analysis, we therefore investigated whether the predictions of the best neural network model explain the variance explained by the linguistic features that modulate inference strength. To this end, we used a slightly simplified\footnote{We removed by-item random intercepts  and by-subject random slopes to facilitate inference. This simplification yielded almost identical estimates as the original model by \newcite{degen2015investigating}.} Bayesian implementation of the mixed-effects model by \newcite{degen2015investigating} that predicted inference strength ratings from hand-mined features. We used the brms \cite{burkner2017brms} and STAN \cite{carpenter2017stan} packages and compared this \textit{original} model to an \textit{extended} model that included both all of the predictors of the original model as well as the the output of the above NN model as a predictor. For this comparison, we investigated whether the magnitude of a predictor in the \textit{original} model significantly decreased in the \textit{extended} model with the NN predictor, based on the reasoning that if the NN predictions explain the variance previously explained by these manually coded predictors, then the original predictor should explain no or less additional variance. 

We approximated the probability that the magnitude of the coefficient for the predictor $i$ ($\beta_i$) in the \textit{extended} model including the NN predictor is smaller than the coefficient in the \textit{original} model, $P(|\beta_i^{extended}| < |\beta_i^{original}|)$, by sampling values for each coefficient from the distributions of the \textit{original} and the \textit{extended} models and comparing the magnitude of the sampled coefficients. We repeated this process 1,000,000 times and treated the simulated proportions as approximate probabilities.

An issue with this analysis is that  estimating the regression model only on the items in the held-out test set yields very wide credible intervals for some of the predictors--in particular for some of the interactions--since the model infers these values from very little data. We therefore performed this regression analysis (and the subsequent analyses) on the entire data. However, while we estimated the regression coefficients from all the data, we crucially obtained the NN predictions through 6-fold cross-validation (without additional tuning of hyperparameters), so that the NN model always made predictions on data that it had not seen during training. This did yield the same qualitative results as the analyses only performed on the held-out test items (see Appendix~\ref{app:regression}) but it also provided us with narrower credible intervals that highlight the differences between the coefficient estimates of the two models.

\figref{fig:cv-coefficients} shows the estimates of the coefficients in the original model and the extended model. We find that the NN predictions explain some or all of the variance originally explained by many of the manually coded linguistic features: the estimated magnitude of the predictors for partitive, determiner strength, linguistic mention, subjecthood, modification, utterance length,  and two of the interaction terms decreased in the extended model. These results provide additional evidence that the NN model indeed learned associations between linguistic features and inference strength rather than only explaining  variance caused by individual items. This is particularly true for the grammatical and lexical features; we find that the NN predictor explains most of the variance originally explained by the partitive, subjecthood, and modification predictors. More surprisingly, the NN predictions also explain a lot of the variance originally explained by the determiner strength predictor, which was empirically determined by probing human interpretation and is not encoded explicitly in the surface form utterance.\footnote{As explained above, \newcite{degen2015investigating} obtained strength ratings by asking participants to rate the similarity of the original utterance and an utterance without the determiner \textit{some (of)}.} One potential explanation for this is that strong and weak \textit{some} have different context distributions. For instance, weak \emph{some} occurs in existential \emph{there} constructions and with individual-level predicates, whereas strong \emph{some} tends not to \cite{milsark1974,mcnally1998,carlson1977}. Since pre-trained word embedding models capture a lot of distributional information, the NN model is presumably able to learn this association. 

\begin{figure*}[t]
	\includegraphics[trim={0 0.5cm 0 0},clip,width=\textwidth]{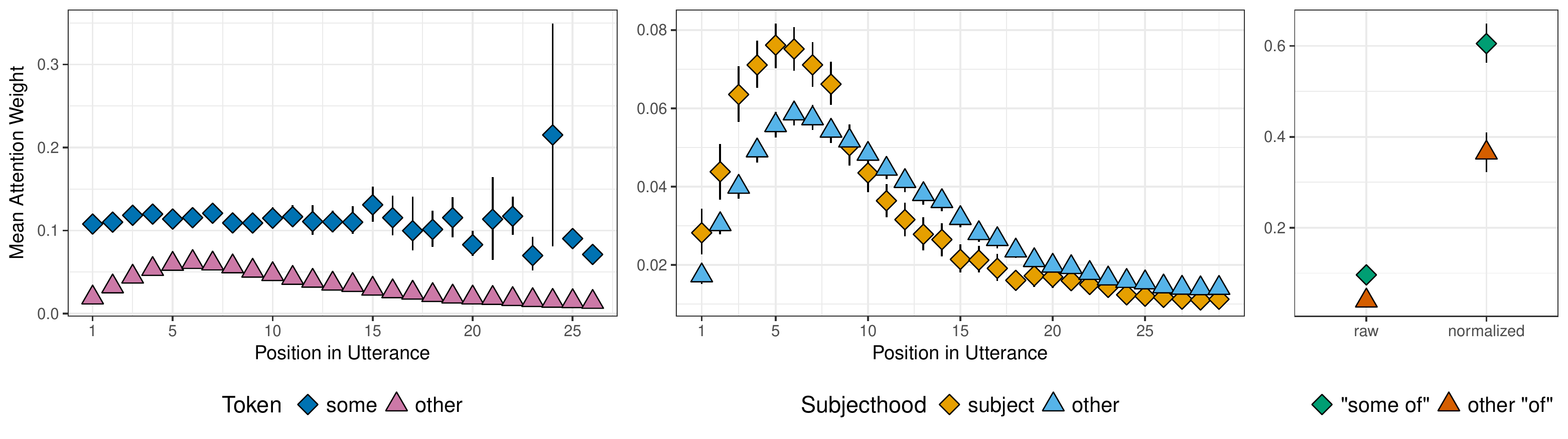}
	\caption{\textbf{Left}: Average attention weights at each token position for \emph{some} and other tokens. \textbf{Center}:  Average attention weights at each token position for utterances with subject and non-subject \emph{some}-NPs. \textbf{Right:}
	Average attention weights of \emph{of}-tokens in partitive \emph{some}-NPs and weights of other \emph{of}-tokens. In the normalized cases, we take only the utterances with multiple \emph{of}-tokens into account and re-normalize the attention weights across all \emph{of}-tokens in one utterance. Error bars indicate 95\% bootstrapped confidence intervals.}
	\label{fig:avgattn}
\end{figure*}

\paragraph{Attention weight analysis.} As a final type of analysis, we analyzed the attention weights that the model used for combining the token embeddings to a sentence embedding. Attention weight analyses have been successfully used for inspecting and debugging model decisions (e.g., \citealp{lee2017interactive,ding2017visualizing,wiegreffe2019attention,vashishth2019attention}; but see \citealp{serrano2019attention}, and \citealp{jain2019attention}, for critical discussions of this approach). Based on these results, we expected the model to attend more to tokens that are relevant for making predictions.\footnote{As pointed out by one of the reviewers, given the transformer architecture,  BERT token representations are influenced by numerous tokens of the input sentence and therefore it could be that the output representation of the $i$-th token ultimately contains very little information about the  $i$-th token that was input to the model. Consequently, it could be that the attention weights do not provide information about which tokens the model attends to. To rule out this possibility, we also conducted the attention weight analysis for the model using static GloVe embeddings, which always exclusively represent the input token, and we found the same qualitative patterns as reported in this section, suggesting that the attention weights provide information about the tokens that are most informative for making predictions. Nevertheless, we do want to caution researchers from blindly trusting attention weight analyses and recommend using this type of analysis only in combination with other types of analyses as we have done in this work.} Given that many of the hand-mined features that predict inference strength occur within or in the vicinity of the \emph{some}-NP, we should therefore expect the model to attend most to the \emph{some}-NP. 

To test this, we first explored whether the model attended on average more to \emph{some} than to other tokens in the same position. Further, we exploited the fact that in English, subjects generally occur early in a sentence. If the model attends to the vicinity of the \emph{some}-NP, the average attention weights should be higher at early positions in utterances with a subject \emph{some}-NP compared to utterances with a non-subject  \emph{some}-NP, and conversely for late utterance positions. We thus compared the average attention weights for each position across utterances with subject versus non-subject \emph{some}-NPs.   To make sure that any effects were not only driven by the attention weight of the \textit{some}-tokens, we set the attention weights of the token corresponding to \textit{some} to $0$ and re-normalized the attention weights for this analysis. Further, since the attention weights are dependent on the number of tokens in the utterance, it is crucial that the average utterance length across the two compared groups be matched. We addressed this by removing outliers and limiting our analysis to utterances up to length 30 (1,028 utterances), which incidentally equalized the number of tokens across the two groups. These exclusions resulted in tiny differences in the average attention weights, but the qualitative patterns are not affected.

The left panel of \figref{fig:avgattn} shows the average attention weight by position for \emph{some} versus other tokens. The model assigns much higher weight to \emph{some}. The center panel of \figref{fig:avgattn} shows the average attention weight by position for subject vs.~non-subject  \emph{some}-NP utterances. The attention weights are generally higher for tokens early in the utterance,\footnote{This is in part an artifact of shorter utterances which distribute the attention weights among fewer tokens.} but the attention weights of utterances with a subject \textit{some}-NP are on average higher for tokens early in the utterance compared to utterances with the \textit{some}-NP in non-subject positions. Both of these findings provide evidence that the model assigns high weight to the tokens within and surrounding the \textit{some}-NP.\footnote{The regression analysis suggests that the model learned to make use of the subjecthood feature and previous work on probing behavior of contextual word representations has found that such models are capable of predicting dependency labels, including subjects \citep[e.g.,][]{liu2019}. We therefore also hypothesize that the representations of tokens that are part of a subject \textit{some}-NP contain information about the subjecthood status. This in return could be an important feature for the output layer of the model and therefore be providing additional signal for the model to attend to these tokens.}

In a more targeted analysis to assess whether the model learned to use the partitive feature, we examined whether the model assigned higher attention to the preposition \emph{of} in partitive \emph{some}-NPs compared to when \emph{of} occurred elsewhere. 
As utterance length was again a potential confound, we conducted the analysis separately on the full set of utterances with raw attention weights and on a subset that included only utterances with at least two instances of \emph{of} (128 utterances), in which we renormalized the weights of \emph{of}-tokens to sum to 1.

Results are shown in the right panel of \figref{fig:avgattn}. The attention weights were higher for \emph{of} tokens in partitive \emph{some}-NPs, suggesting that the model learned an association between partitive \emph{of} in \emph{some}-NPs and inference strength.

\section{Context, revisited}
\label{sec:context}

In the tuning experiments above, we found that including the preceding conversational context in the input to the model did not improve or  lowered prediction accuracy.\footnote{As suggested by a reviewer, we conducted post-hoc experiments in which we limited the conversational context to the preceding 2 or 5 utterances, which presumably have a higher signal-to-noise ratio than a larger conversational context of 10 preceding utterances. In these experiments, we again found that including the conversational context did not improve model predictions.} At the same time, we found that the model is capable of making accurate predictions in most cases without taking the preceding context into account. Taken together, these results suggest either that the conversational context is not necessary and one can draw inferences from the target utterance alone, or that the model does not make adequate use of the preceding context. 

\citet{degen2015investigating} did not systematically investigate whether the preceding conversational context was used by participants judging inference strength. To assess the extent to which the preceding context in this dataset affects inference strength, we re-ran her experiment, but without presenting participants with the preceding conversational context. We recruited 680 participants on Mechanical Turk who each judged 20 or 22 items, yielding 10 judgments per item. If the context is irrelevant for drawing inferences, then mean inference strength ratings should be very similar across the two experiments, suggesting the model may have rightly learned to not utilize the context. If the presence of context affects inference strength,  ratings should differ across experiments, suggesting that the model's method of integrating context is ill-suited to the task.

The new, no-context ratings correlated with the original ratings ($r=0.68$, see Appendix~\ref{app:no-context}) but were overall more concentrated towards the center of the scale, suggesting that in many cases, participants who lacked information about the conversational context were unsure about the strength of the scalar inference. Since the original dataset exhibited more of a bi-modal distribution with fewer ratings at the center of the scale, this suggests that the broader conversational context contains important cues to scalar inferences.

For our model, these results suggest that the representation of the conversational context is inadequate, which highlights the need for more sophisticated representations of linguistic contexts beyond the target utterance.\footnote{The representation of larger linguistic context is also important for span-based question-answer (QA) systems (e.g., \citealp{hermann2015,chen2018,devlin2019bert}) and adapting methods from QA to predicting scalar inferences would be a promising extension of the current model.} We further find that the model trained on the original dataset is worse at predicting the no-context ratings ($r=0.66$) than the original ratings ($r=0.78$), which is not surprising considering the imperfect correlation between ratings across experiments, but also provides additional evidence that participants indeed behaved differently in the two experiments. 

\section{Conclusion and future work}

We showed that despite lacking specific pragmatic reasoning abilities, neural network-based sentence encoders are capable of harnessing the linguistic signal to learn to predict human inference strength ratings from \emph{some} to \emph{not all} with high accuracy. Further, several model behavior analyses provided consistent evidence that the model learned associations between previously established linguistic features and the strength of scalar inferences. 

In an analysis of the contribution of the conversational context, we found that humans make use of the preceding context whereas the models we considered failed to do so adequately. Considering the importance of context in drawing both scalar and other inferences in communication \cite{Grice1975,clark1992arenas,bonnefon2009some,Zondervan2010,bergengrodner2012,goodman2013knowledge,degen2015wonky}, the development of  appropriate representations of larger context is an exciting avenue for future research. 

We also only considered the supervised setting in which the model was trained to predict inference strength. It would be interesting to investigate how much supervision is necessary and, for example, to what extent a model trained to perform another task such as predicting natural language inferences is able to predict scalar inferences (see \newcite{Jiang2019b} for such an evaluation of predicting speaker commitment, and \newcite{Jeretic2020} for an evaluation of different NLI models for predicting lexically triggered scalar inferences). 

One further interesting line of research would be to extend this work to other pragmatic inferences. Recent experimental work has shown that inference strength is variable across scale and inference type \citep{doran2012novel, van2016scalar}.  We treated \emph{some} as a case study in this work, but none of our modeling decisions are specific to \emph{some}. It would be straightforward to train similar models for other types of inferences.

Lastly, the fact that the attention weights  provided insights into the model's decisions suggests possibilities for using neural network models for developing more precise theories of pragmatic language use. Our goal here was to investigate whether neural networks can learn associations for already established linguistic features but it would be equally interesting to investigate whether such models could be used to discover new features, which could then be verified in experimental and corpus work, potentially providing a model-driven approach to experimental and formal pragmatics.

\section*{Acknowledgements}
We thank the anonymous reviewers for their
thoughtful feedback. 
We also gratefully acknowledge Leyla Kursat for collecting the 
no-context inference strength ratings, and we thank Jesse Mu, Shyamal Buch, 
Peng Qi, Marie-Catherine de Marneffe, Tal Linzen, and the members of the 
ALPS lab and the JHU Computational Psycholinguistics group for helpful discussions.

\bibliography{naaclhlt2019}
\bibliographystyle{acl_natbib}

\clearpage

\appendix

\onecolumn

\section{Hyperparameter tuning}
\label{app:hyperparameter} 
\begin{figure}[h]
\centering
	\includegraphics[width=\textwidth,trim={0 0.4cm 0 0},clip]{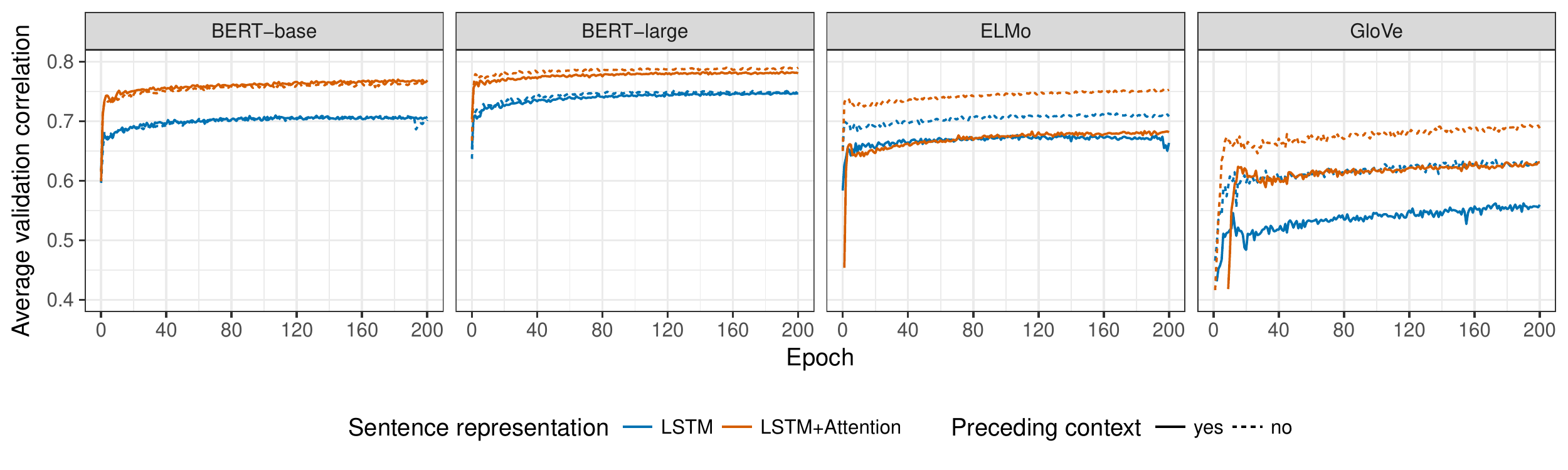}
	\caption{Correlation between each model's predictions on valuation set and empirical means, by training epoch.}
	\label{fig:learning-curves}
\end{figure}

\figref{fig:learning-curves} shows the  learning curves averaged over the 5 cross-validation tuning runs for models using different word embeddings. As these plots show, the attention layer improves predictions; contextual word embeddings lead to better results than the static GloVe embeddings; and including the conversational context does not improve predictions and in some cases even lowers prediction accuracy.

\section{Regression analysis on held-out test data}
\label{app:regression}
\begin{figure}[h]
\center
\includegraphics[width=\textwidth]{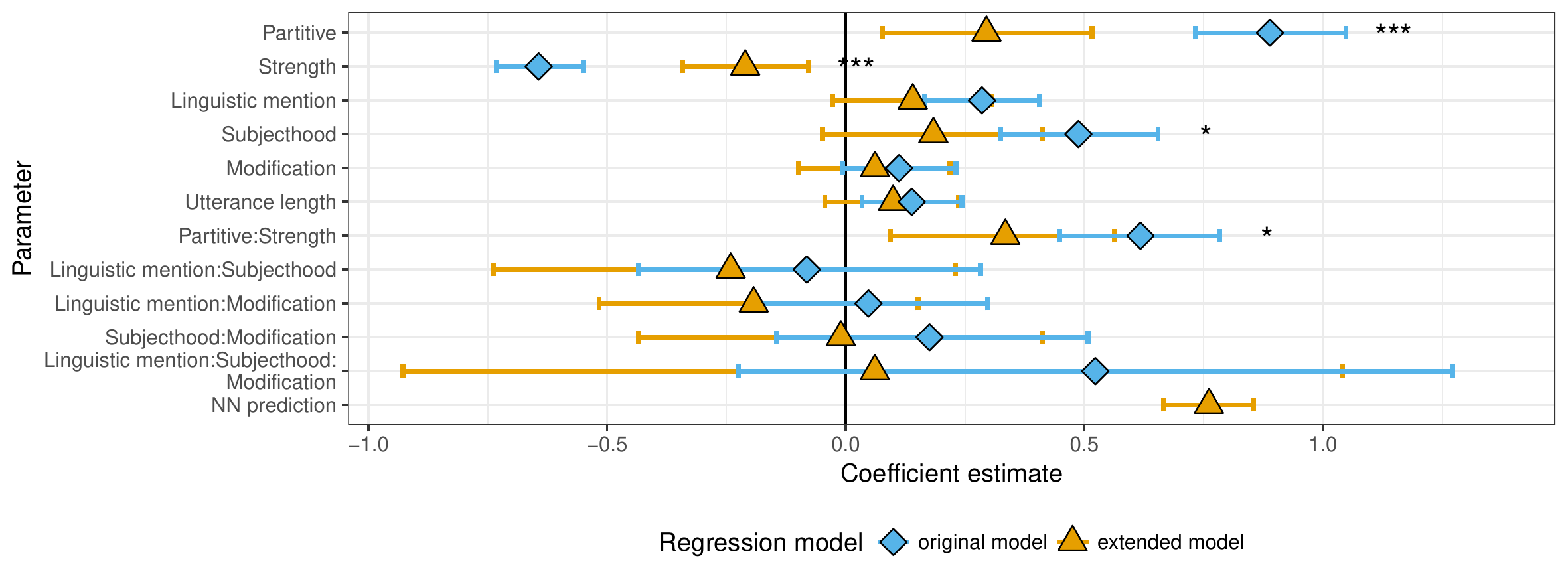}
\caption{Maximum a posteriori estimates and 95\%-credible intervals of coefficients for original and extended Bayesian mixed-effects regression models predicting the inference strength ratings on the held-out test set. */**/*** indicate that the probability of the coefficient of the original model having a larger magnitude than the coefficient of the extended model is less than 0.05, 0.01, and 0.001, respectively.}
	\label{fig:eval-coefficients}
\end{figure}

\noindent\figref{fig:eval-coefficients} shows the estimates of the predictors in the original and extended Bayesian mixed-effects models estimated only on the held-out test data. We find the same qualitative effects as in \figref{fig:cv-coefficients}, but since these models were estimated on much less data (only 408 items), there is a lot of uncertainty in the estimates and therefore quantitative comparisons between the coefficients of the different models are less informative.

\clearpage

\section{List of manually constructed sentences}
\label{app:sentences}
Tables~\ref{tbl:artificial-sentences-1} and \ref{tbl:artificial-sentences-2} show the 25 manually created sentences for the analyses described in the minimal pairs analysis in \sectionref{sec:analyses}. As described in the main text, we created 16 variants of the sentence with the \emph{some}-NP in subject position (sentences in the left column), and 16 variants of the sentence with the \emph{some}-NP in object position (sentences in the right column), yielding in total 800 examples.

\begin{table}[h]
    \centering
    \begin{tabular}{|p{.49\columnwidth}|p{.49\columnwidth}|}
        \hline
        Some of the attentive waiters at the gallery opening poured the white wine that my friend really likes. & The attentive waiters at the gallery opening poured some of the white wine that my friend really likes.  \\
        \hline
        Some of the experienced lawyers in the firm negotiated the important terms of the acquisition. & The experienced lawyers in the firm negotiated some of the important terms of the acquisition. \\
        \hline
        Some of the award-winning chefs at the sushi restaurant cut the red salmon from Alaska. & The award-winning chefs at the sushi restaurant cut some of the red salmon from Alaska. \\
        \hline
        Some of the brave soldiers who were conducting the midnight raid warned the decorated generals who had served in a previous battle. & The brave soldiers who were conducting the midnight raid warned some of the decorated generals who had served in a previous battle. \\
        \hline
        Some of the eccentric scholars from the local college returned the old books written by Camus. & The eccentric scholars from the local college returned some of the old books written by Camus. \\
        \hline
        Some of the entertaining magicians with top hats shuffled the black cards with dots. & The entertaining magicians with top hats shuffled some of the black cards with dots. \\
        \hline
        Some of the convicted doctors from New York called the former patients with epilepsy. & The convicted doctors from New York called some of the former patients with epilepsy. \\
        \hline
        Some of the popular artists with multiple albums performed the fast songs from their first album. & The popular artists with multiple albums performed some of the fast songs from their first album. \\
        \hline
        Some of the angry senators from red states impeached the corrupt presidents from the Republican party. & The angry senators from red states impeached some of the corrupt presidents from the Republican party. \\
        \hline
        Some of the underfunded researchers without permanent employment transcribed the recorded conversations that they collected while doing fieldwork.  & The underfunded researchers without permanent employment transcribed some of the recorded conversations that they collected while doing fieldwork. \\
        \hline
        Some of the sharp psychoanalysts in training hypnotized the young clients with depression. & The sharp psychoanalysts in training hypnotized some of the young clients with depression. \\
        \hline
        Some of the harsh critics from the Washington Post read the early chapters of the novel. & The harsh critics from the Washington Post read some of the early chapters of the novel. \\
        \hline
       Some of the organic farmers in the mountains milked the brown goats who graze on the meadows. & The organic farmers in the mountains milked some of the brown goats who graze on the meadows. \\
        \hline
        Some of the artisanal bakers who completed an apprenticeship in France kneaded the gluten-free dough made out of spelt. &  The artisanal bakers who completed an apprenticeship in France kneaded some of the gluten-free dough made out of spelt. \\ 
        \hline
        Some of the violent inmates in the high-security prison reported the sleazy guards with a history of rule violations. & The violent inmates in the high-security prison reported some of the sleazy guards with a history of rule violations. \\
       
        \hline
       
    \end{tabular}
    \caption{Manually constructed sentences used in the minimal pair analyses. Sentences in the left column have a \emph{some}-NP in subject position; sentences on the right have a \emph{some}-NP object position.}
    \label{tbl:artificial-sentences-1}
\end{table}

\begin{table}[h]
    \centering
    \begin{tabular}{|p{.49\columnwidth}|p{.49\columnwidth}|}
     \hline
        Some of the eager managers in the company instructed the hard-working sales representatives in the steel division about the new project management tool. & The eager managers in the company instructed some of the hard-working sales representatives in the steel division about the new project management tool. \\
     \hline
     
        Some of the brilliant chemists in the lab oxidized the shiny metals extracted from ores. & The brilliant chemists in the lab oxidized some of the shiny metals extracted from ores. \\
        \hline
        Some of the adventurous pirates on the boat found the valuable treasure that had been buried in the sand. & The adventurous pirates on the boat found some of the valuable treasure that had been buried in the sand. \\
        \hline
         Some of the mischievous con artists at the casino tricked the elderly residents of the retirement home. & The mischievous con artists at the casino tricked some of the elderly residents of the retirement home. \\
        \hline
        Some of the persistent recruiters at the conference hired the smart graduate students who just started a PhD as interns. & The persistent recruiters at the conference hired some of the smart graduate students who just started a PhD as interns.  \\
        \hline
        Some of the established professors in the department supported the controversial petitions that were drafted by the student union. & The established professors in the department supported some of the controversial petitions that were drafted by the student union.
        \\
        \hline
        Some of the muscular movers that were hired by the startup loaded the adjustable standing desks made out of oak onto the truck. & The muscular movers that were hired by the startup loaded some of the adjustable standing desks made out of oak onto the truck. \\
        \hline
        Some of the careful secretaries at the headquarter mailed the confidential envelopes with the bank statements. & The careful secretaries at the headquarter mailed some of the confidential envelopes with the bank statements. \\
        \hline
        Some of the international stations in South America televised the early games of the soccer cup. & The international stations in South America televised some of the early games of the soccer cup. \\
        \hline
        Some of the wealthy investors of the fund excessively remunerated the successful brokers working at the large bank. & The wealthy investors of the fund excessively remunerated some of the successful brokers working at the large bank. \\
        \hline

    \end{tabular}
    \caption{Manually constructed sentences used in the minimal pair analyses (continued).}
    \label{tbl:artificial-sentences-2}
    \vspace{5em}
\end{table}

\clearpage

\section{Results from no-context experiment}
\label{app:no-context} 
\begin{figure}[h]
\centering
	\includegraphics[width=.49\textwidth]{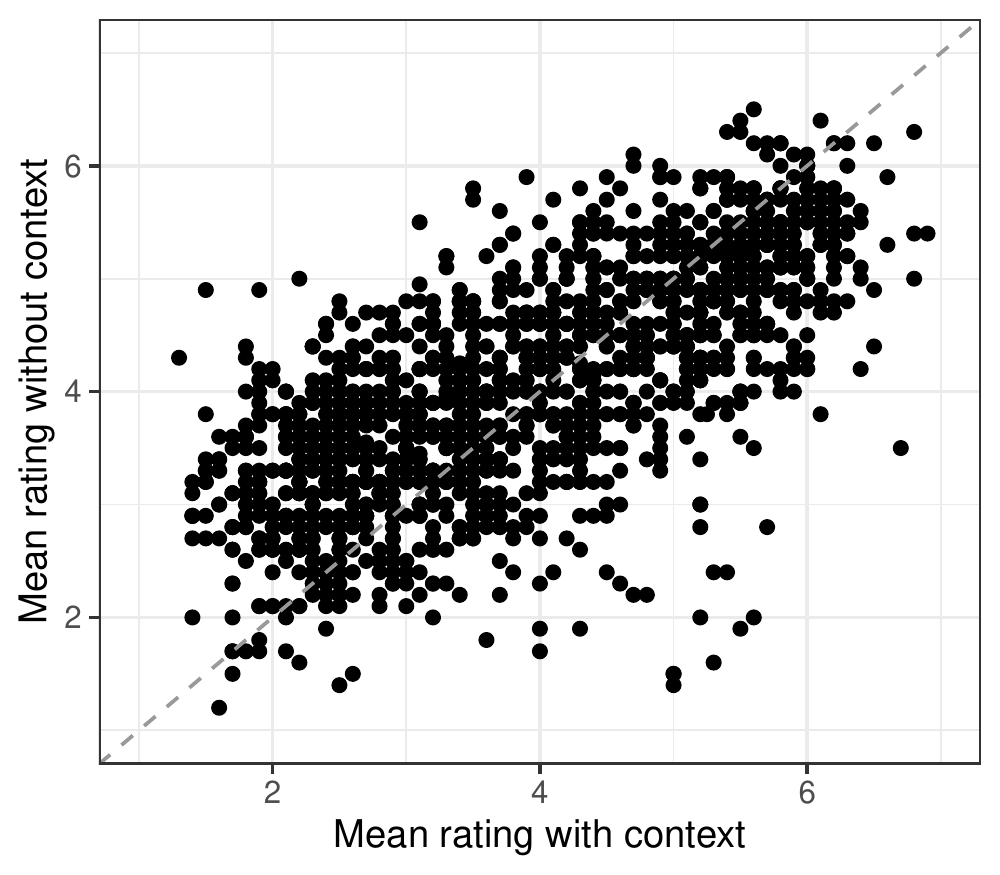}
	\caption{Mean inference strength ratings for items without context (new) against items with context (original), $r=.68$.}
	\label{fig:context-nocontext}
\end{figure}

\figref{fig:context-nocontext} shows the correlation between the mean inference strength ratings for each item in the experiment from \citet{degen2015investigating} and the mean strength ratings from the new no-context experiment, discussed in \sectionref{sec:context}.

\end{document}